\documentclass{article}

\usepackage[final]{svrhm_2019}

\usepackage[utf8]{inputenc} 
\usepackage[T1]{fontenc}    
\usepackage{hyperref}       
\usepackage{url}            
\usepackage{booktabs}       
\usepackage{amsfonts}       
\usepackage{nicefrac}       
\usepackage{microtype}      

\usepackage{graphicx}       

\title{Shared Visual Abstractions}

\author{%
  Tom White \\
  School of Design\\Victoria University of Wellington\\Wellington, New Zealand \\
  \texttt{tom.white@vuw.ac.nz} \\
}

\begin{document}

\maketitle

\begin{abstract}
This paper presents abstract art created by neural networks and broadly recognizable across various computer vision systems. The existence of abstract forms that trigger specific labels independent of neural architecture or training set suggests convolutional neural networks build shared visual representations for the categories they understand. Computer vision classifiers encountering these drawings often respond with strong responses for specific labels - in extreme cases stronger than all examples from the validation set. By surveying human subjects we confirm that these abstract artworks are also broadly recognizable by people, suggesting visual representations triggered by these drawings are shared across human and computer vision systems.
\end{abstract}

\begin{figure}[!htb]
  \centering
  \includegraphics[width=14cm]{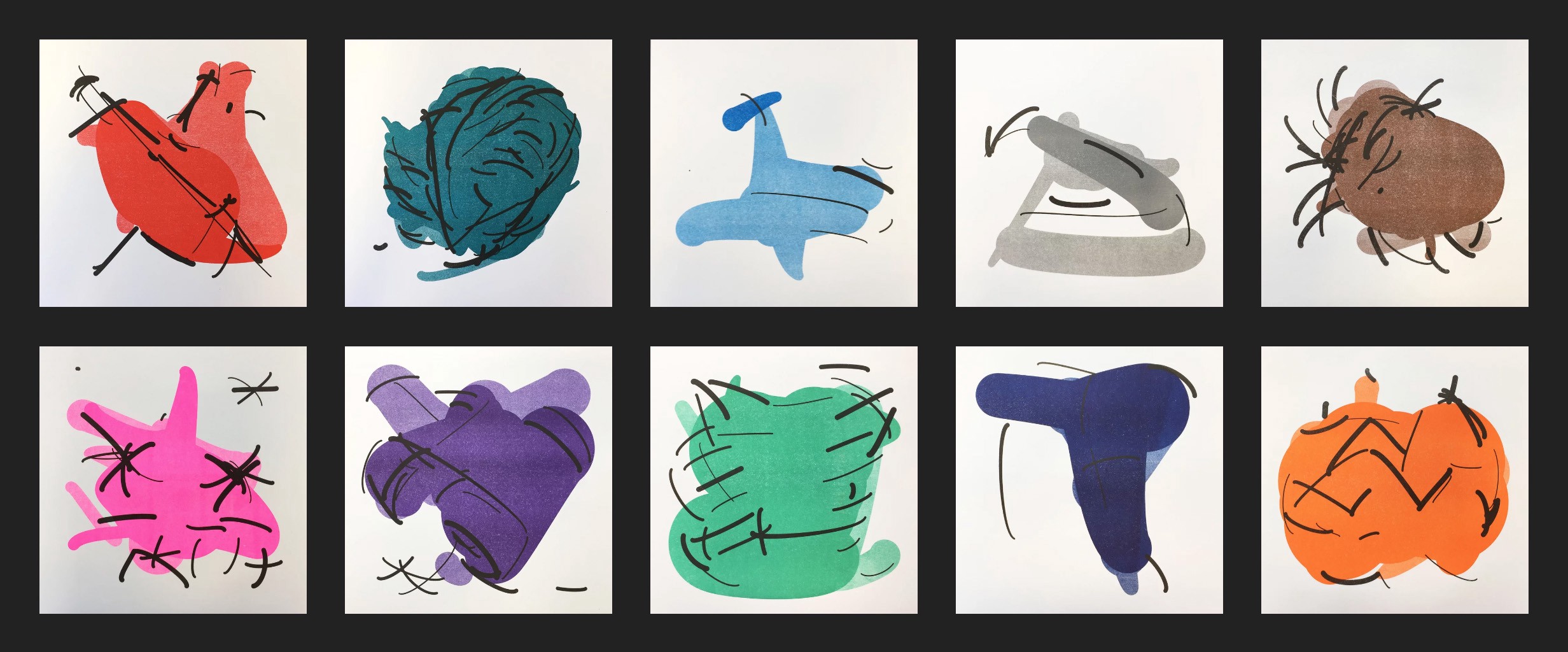}
  \caption{
Ink prints demonstrating visual abstractions created from ImageNet categories: cello, cabbage, hammerhead shark, iron, tick, starfish, binoculars, measuring cup, blow dryer, and jack-o-lantern. These are generally recognizable by ImageNet trained neural networks independent of neural architecture.
  }
\end{figure}

\section{Introduction}

Visual abstraction is central to written human communication. The earliest known writing system dates to 3000 BC (figure 2); cuneiform used simplified drawn pictures to record quantities of grain, sheep and cattle, and over millennia these impressions made on damp clay were standardized using a wedge shaped writing instrument \citep*{historicwriting}. The earliest cuneiform symbols were visually abstract forms meant to represent categories and the shapes pressed into clay were streamlined to contain only the most discriminative features relative to other categories.

\begin{figure}
    \centering
    \begin{minipage}{0.375\textwidth}
        \centering
        \includegraphics[width=0.9\textwidth]{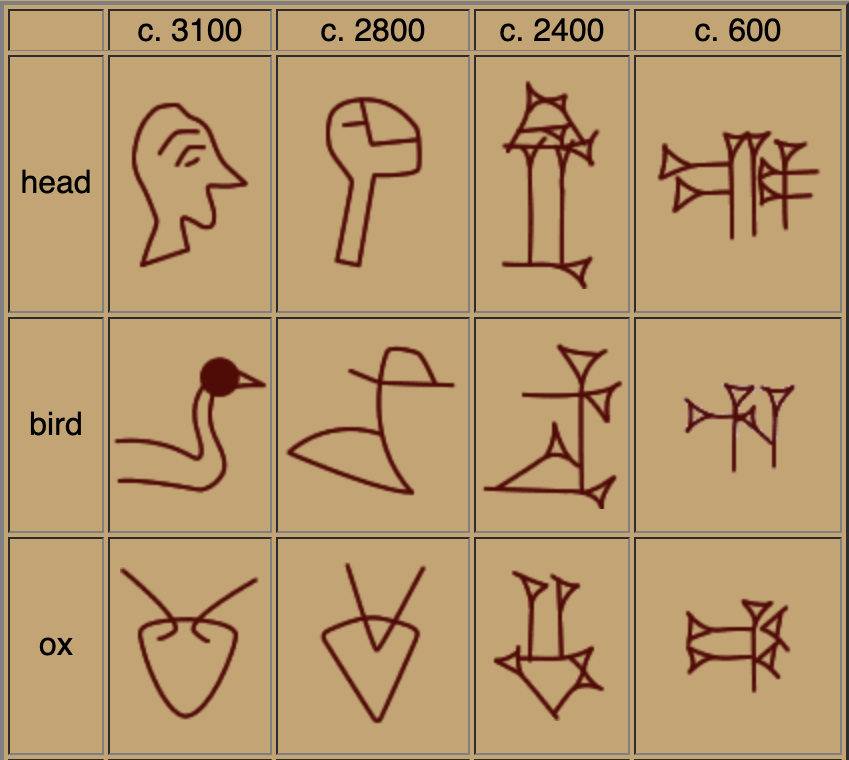} 
        \caption{
Cuneiform symbols for head, bird, and ox showing how each originated in a visual abstraction for the named concept (from \citet{cunwriting})
        }
    \end{minipage}\hfill
    \begin{minipage}{0.56\textwidth}
        \centering
        \includegraphics[width=0.9\textwidth]{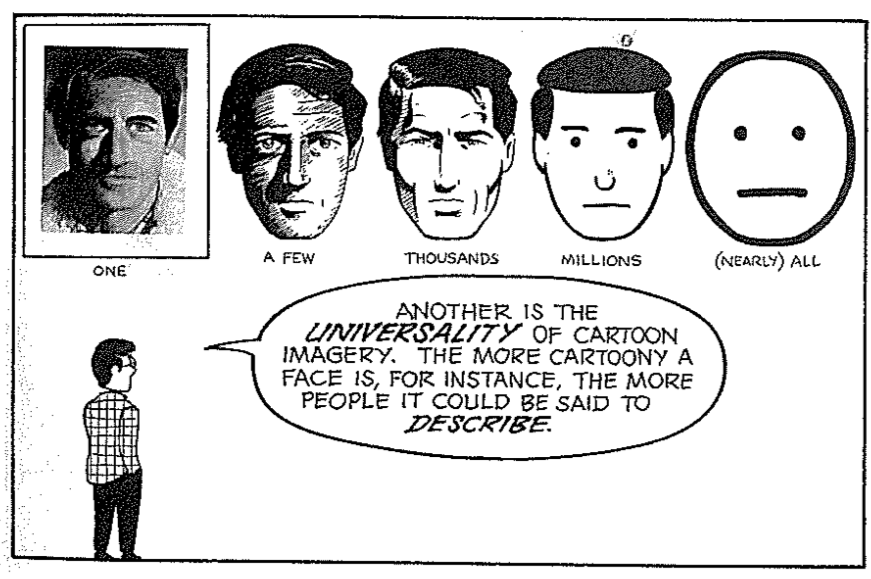} 
        \caption{
The concept of “amplification through simplification” from \citet{McCloud93aa} explains the universality of cartoon imagery and here visually demonstrates how the number discriminative features in a visual abstraction can result in overfitting or underfitting the intended target class.
        }
    \end{minipage}
\end{figure}

In an analysis of comics, \citet{McCloud93aa} examines various levels of visual abstraction and describes cartooning as a form of “amplification through simplification” (figure 3). By eliminating features that are specific to individuals or even modes of a visual category, a cartoonist can incrementally strip down an image in order to amplify meaning by making the concept more universal. Discriminative visual features can be introduced or removed in order to balance overfitting or underfitting an image relative to the intended target concept.

Adversarial examples research provides techniques for generating imagery to trigger specific categories in machine learning models \citep{advex}. These techniques are generally discussed in the context of security and reliability, and so are constrained to small and presumably imperceptible perturbations of a source image. \citet{ilyas2019adversarial} suggests that adversarial examples are simply utilizing unintended predictive features of the dataset which diverge from those used in human classification.

\section{Visual Abstractions}

This work investigates the possibility that neural network image classifiers are also capable of creating visual abstractions generally recognizable by other computer vision systems and perhaps humans as well. A perception driven drawing system was created to allow neural network classifiers to draw ink and paper abstract shapes that maximize class based responses. \citet{perceptionengines} covers this drawing system in more detail and a reference implementation is available online\footnote{\url{https://github.com/dribnet/perceptionengines}}. 

Inspired by adversarial examples results demonstrating transferability \citep{transferspace}, the drawing system uses an ensemble of trained networks to evaluate drawings which can be executed as ink prints and searches the space of possibilities using random search \citep{randsearch}. However, unlike adversarial examples the images being tested are not constrained to small perturbations - they are instead constrained only by the degrees of freedom provided in the drawing system itself.

This system was initially tested in the context of high accuracy pre-trained ImageNet (ILSVRC) classifiers such as InceptionV3 \citep{SzegedyVISW15} and a single drawing style. Varying only the stroke placement, thickness, and ink color, this system was able to produce several abstract ink drawings with strong responses across different ImageNet classes (figure 1). A photo of each physical print is used to measure the classification accuracy on standard pretrained ImageNet classifiers including several models that were not part of the original ensemble. This testing reveals strong transferability to other pretrained classifiers indicating that these results generalize to other network architectures (figure 4).

\pagebreak

\begin{figure}
  \centering
  \includegraphics[width=13cm]{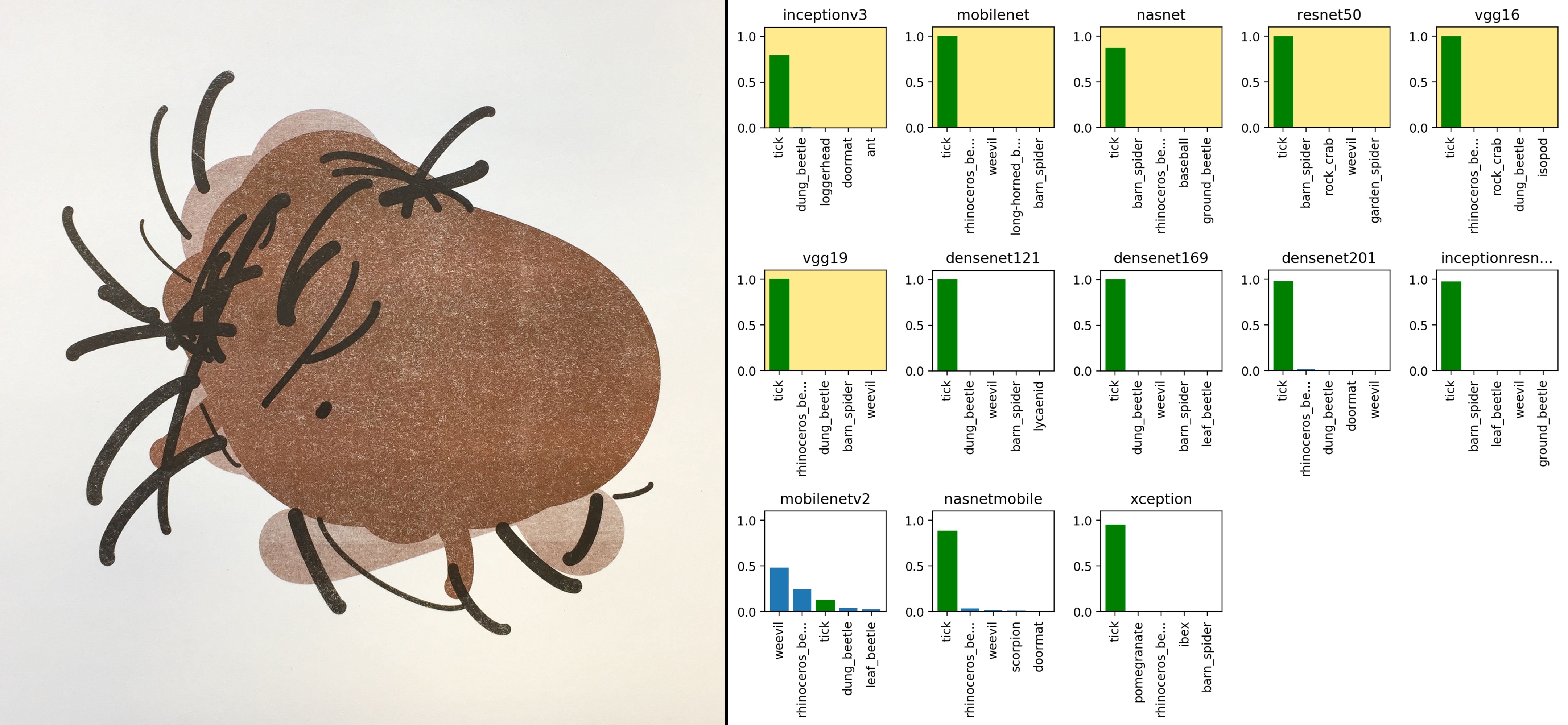}
  \caption{A photo of the tick print yields very high scores for the tick label across 13 tested ImageNet pre-trained models. Each graph shows the top-5 category responses which are generally dominated by the "tick" score shown in green. The graphs with yellow background are from the ensemble of six models used to create this print and high scores when testing on the other seven trained networks suggest that this result generalizes well across other architectures.
  }
\end{figure}

These results indicate that neural networks are capable of producing visual abstractions shared broadly across classifiers trained on the same datasets. The response strength on these simplified shapes was often stronger than the more detailed in-class example images, reminiscent of McCloud’s description of “amplification through simplification”. This can be verified empirically by comparing the label response strength of the drawing and original validation data for the targeted class (figure 5).

\begin{figure}
  \centering
  \includegraphics[width=13.5cm]{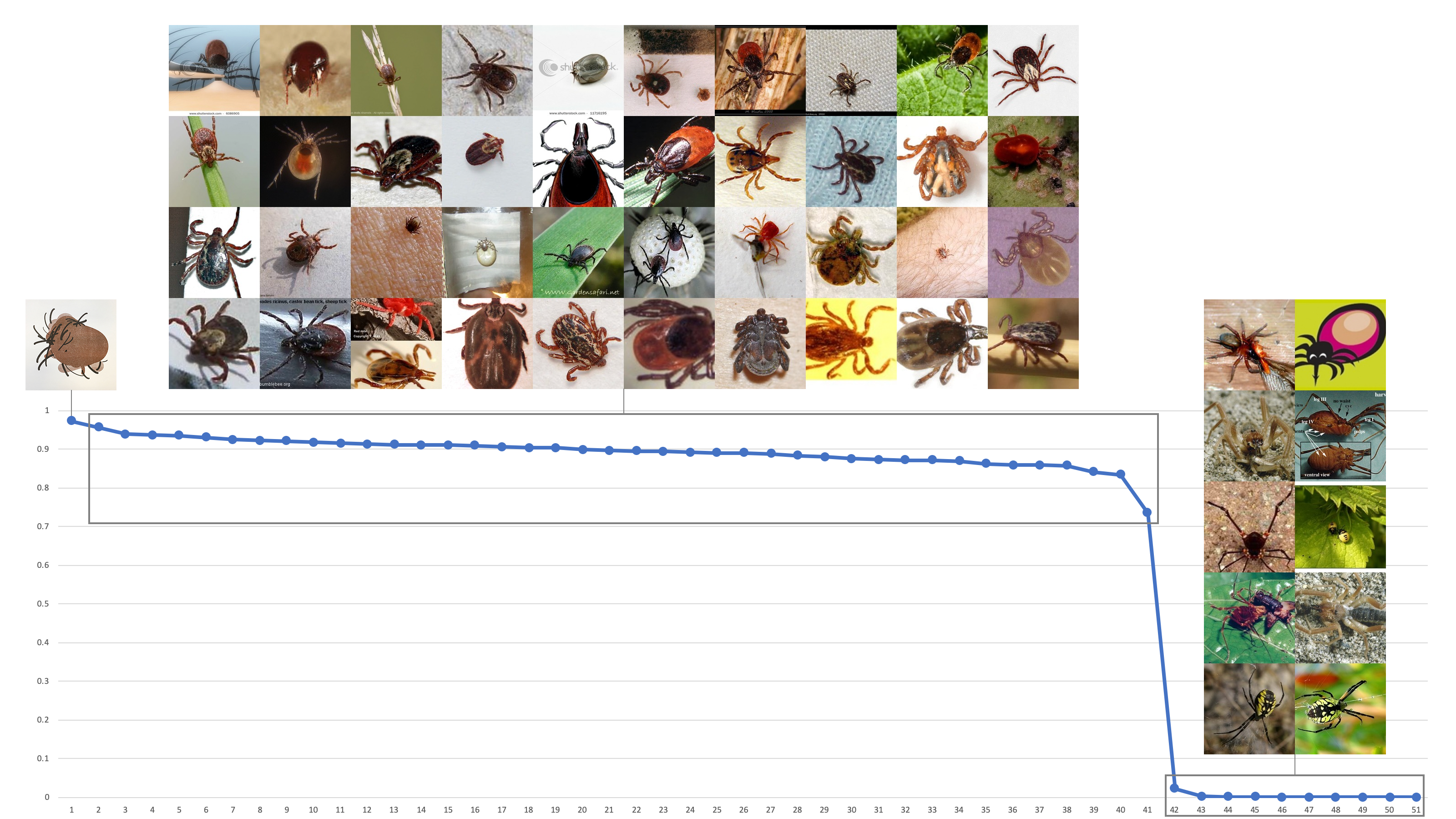}
  \caption{Tick print compared with all 50 ImageNet tick validation images here shown in high and low scoring subgroups. All images are sorted in order of response strength for the tick label on a pretrained InceptionResNetV2 network (y-axis). The tick print elicits stronger responses than all in-class validation images even though this network was not involved in the construction of this print. This strong generalization result suggests that this visual abstraction could be framed as a form of “amplification through simplification” in computer vision systems.}
\end{figure}

\pagebreak

When viewed as a form of adversarial attack that is allowed arbitrary perturbations, these drawings are effective across a much larger class of threat models than usually considered. Adversarial attacks are generally categorized as either white-box attacks requiring the adversary to have complete knowledge of the target model or black-box attacks requiring only queries to the target model that may return complete or partial information \citep{SimpleBlack}. In addition to these we propose a new broader category of a "whiteboard attack" in which the attacker targets a completely unknown model including those on architectures that may not yet exist. These drawings can then be considered whiteboard attacks using only access to the training data and potentially effective even on future neural architectures.

Further work shows that the visual abstractions produced generalize not only across various neural architectures, but also across networks with different training sets. By training a variety of architectures on a custom dataset, target labels on another trained neural network can be targeted via transferability. A recent drawing using this technique was constructed using a custom dataset of over a thousand pictures of human eyes. When shown to commercial image APIs such a Google Vision where the training data is unknown, this drawing elicits label responses including “Face”, “Eye”, “Eyebrow”, and “Eyelash” (figure 6).

\begin{figure}
  \centering
  \includegraphics[width=14cm]{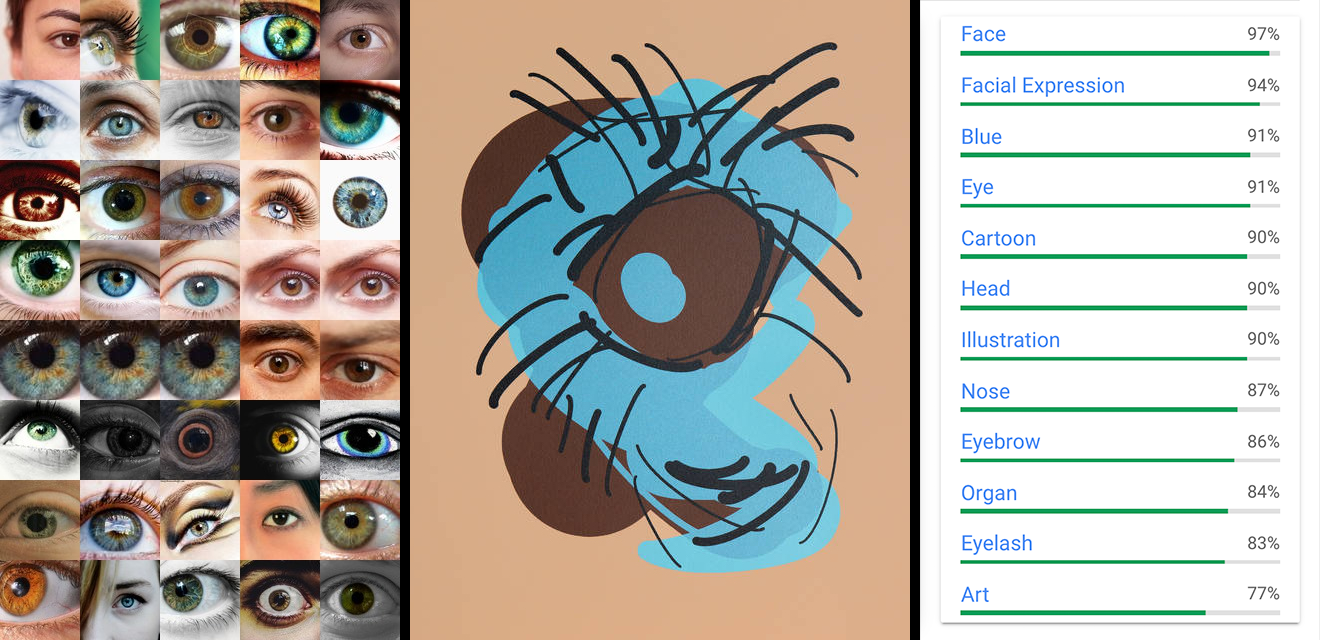}
  \caption{A dataset of thousands of photos of human eyes and faces (left) was used to train neural network classifiers in order to construct a visual abstraction print (center). When this abstract drawing was later shown to the Google Vision API (right), it suggested labels including “Face”, “Eye”, “Eyebrow”, and “Eyelash”.}
\end{figure}

\section{Shared Representations}

These neural network produced drawings have been reported to be evocative of familiar objects when exhibited publicly. To test human interpretability, a multiple-choice survey was made and twelve students unfamiliar with this research were asked to match a label with a pair of drawings created across four different ImageNet categories (see Appendix). Each pair of drawings are visual abstractions created by networks optimizing the ink colours and stroke placement. In this testing all students were able to successfully match all drawings with the correct ImageNet category.

This result indicates that visual abstractions created by convolutional neural networks generalize not only to other network architectures, but also to humans as well. This also provides support that reports of these artworks resembling abstract representations of real world objects may be caused by visual representations shared across humans and machines which are being triggered by these abstract forms.

\pagebreak

\section{Conclusion}

Neural networks trained on real world images are capable of producing visual abstractions of labelled classes. Provided with a special drawing system, these visual abstractions can be recorded and tested to verify that they broadly generalize to other neural network architectures, including those trained on unknown datasets. The responses elicited by these simplified shapes in vision systems often are amplified relative to expected response strengths seen in typical real world data of the same target class. Surveying human subjects establishes that people can reliably predict a training label when shown a neural network created visual abstraction, suggesting the interpretations of these visual abstractions are shared between humans and machines.

\raggedbottom

\bibliographystyle{chicago}
\bibliography{refs}

\pagebreak

\normalsize

\section*{Appendix: Survey}

A survey was given in which four pairs of images are presented. Below each pair of images is the same multiple-choice question:


\fbox{%
\begin{minipage}{33em}

Both of these shapes are intended to represent a type of real world object.

Choose the word that best matches the type of object you think this might be.

Chicken / Tripod / Cabbage / Starfish
\end{minipage}
}

When this survey was presented to 12 students, all students were able to correctly pair the images with the ImageNet class that produced it. Complete survey is shown below.

\begin{figure}[ht!]
  \centering
  \includegraphics[width=14cm]{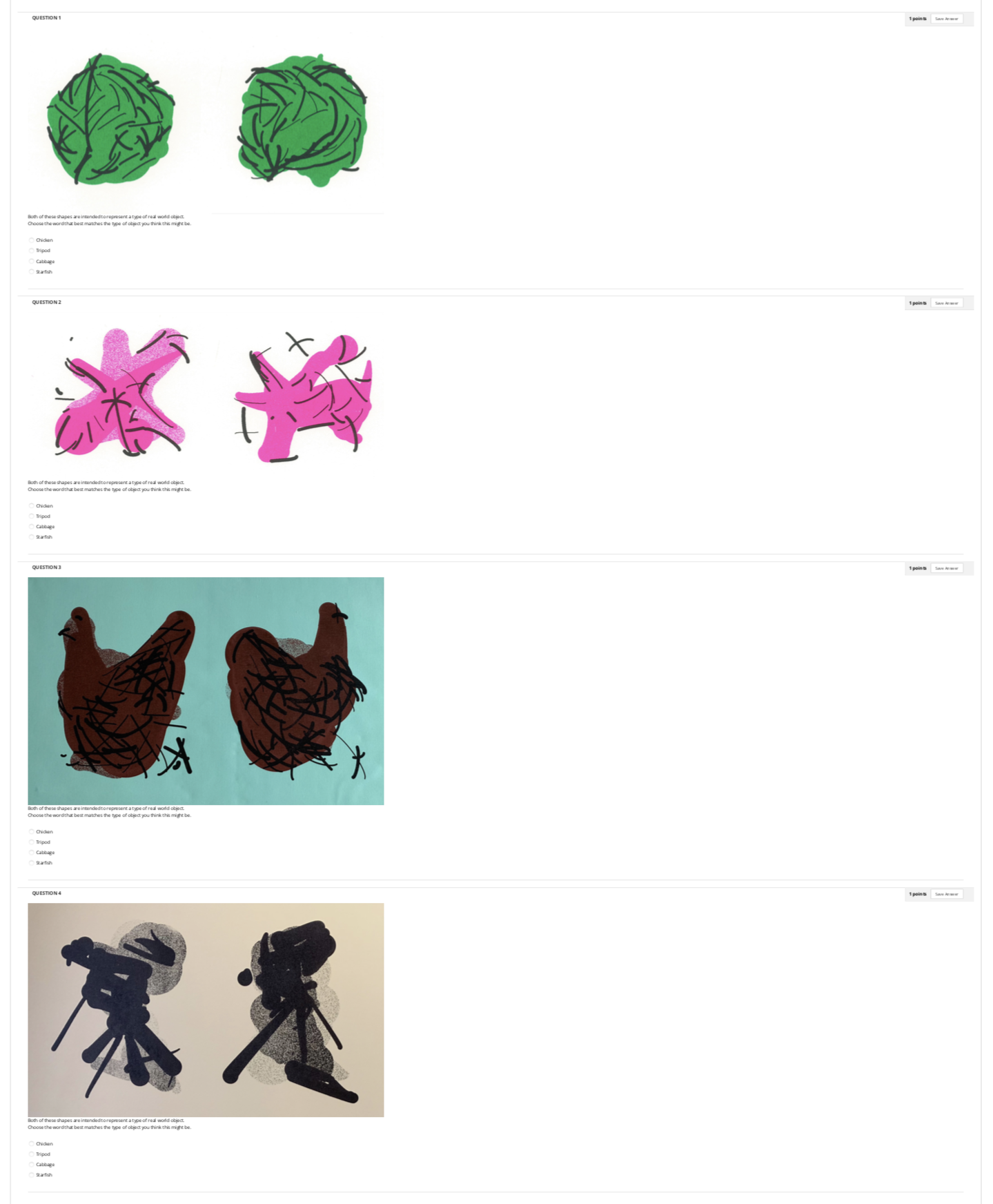}
  \caption{
Survey.
  }
\end{figure}

\end{document}